\documentclass[journal]{IEEEtran}

\usepackage[OT1]{fontenc} 
\usepackage{graphicx}
\usepackage{epstopdf}
\usepackage[tight]{subfigure}
\usepackage[cmex10]{amsmath}
\usepackage{amssymb,amsthm}
\usepackage{array}
\usepackage{booktabs} 
\usepackage{bookmark}
\usepackage{adjustbox}
\usepackage{xcolor}
\usepackage{cite}
\usepackage{pifont}
\usepackage[normalem]{ulem}
\usepackage{multirow}
\usepackage{tikz}

\begin{document}
\title{Adversarial Example in Remote Sensing Image Recognition
\thanks{This work was supported by the National Science Foundation of China (xxxxxxxxxxx,xxxxxxxxxxx,xxxxxxxxxxx)}}

\author{
    Li~Chen,
    Guowei~Zhu,
    Qi~Li,
    Haifeng~Li,~\IEEEmembership{Member,~IEEE}
    \thanks{Li Chen, Guowei Zhu and Haifeng Li are with the School of Geosciences and Info-Physics, 
    Central South University, South Lushan Road, Changsha, 410083, China.
    Email: \mbox{vchenlil@csu.edu.cn}, \mbox{0106160226@csu.edu.cn}, \mbox{lihaifeng@csu.edu.cn}.}
    \thanks{Qi Li is with the School of Computer Science and Engineering,
    Central South University, South Lushan Road, Changsha, 410083, China.
    Email: \mbox{dsjliqi@csu.edu.cn}.}
}

\maketitle

\begin{abstract}
    With the wide application of remote sensing technology in various fields, the accuracy and security requirements for remote sensing images (RSIs) recognition are also increasing. 
    In recent years, due to the rapid development of deep learning in the field of image recognition, RSI recognition models based on deep convolution neural networks (CNNs) outperform traditional hand-craft feature techniques. 
    However, CNNs also pose security issues when they show their capability of accurate classification. By adding a very small variation of the adversarial perturbation to the input image, the CNN model can be caused to produce erroneous results with extremely high confidence, and the modification of the image is not perceived by the human eye. This added adversarial perturbation image is called an adversarial example, which poses a serious security problem for systems based on CNN model recognition results. 
    This paper, for the first time, analyzes adversarial example problem of RSI recognition under CNN models. In the experiments, we used different attack algorithms to fool multiple high-accuracy RSI recognition models trained on multiple RSI datasets. 
    The results show that RSI recognition models are also vulnerable to adversarial examples, and the models with different structures trained on the same RSI dataset also have different vulnerabilities. For each RSI dataset, the number of features also affects the vulnerability of the model. Many features are good for defensive adversarial examples.
    Further, we find that the attacked class of RSI has an attack selectivity property.
    The misclassification of adversarial examples of the RSIs are related to the similarity of the original classes in the CNN feature space. 
    In addition, adversarial examples in RSI recognition are of great significance for the security of remote sensing applications, showing a huge potential for future research.
\end{abstract}


\IEEEpeerreviewmaketitle

\section{Introduction}

\IEEEPARstart{W}{ith} the substantial progress of remote sensing technology, the automatic interpretation of remote sensing images (RSIs) has greatly improved~\cite{lillesand2015remote}. RSIs with higher resolution lead to the wide use of RSI recognition system in crop classification~\cite{kussul2017deep}, forest resource survey~\cite{yuan2015survey}, land cover classification~\cite{scott2017training}, building detection and other fields~\cite{manno2015orientation}. The automatic interpretation of RSIs with robust and high accuracy can bring huge economic efficiency.

The performance of image recognition, which is the first step in RSI recognition system, is especially important. A good image recognition algorithm can extract target features effectively and quickly locate new targets. In the past, we first used feature extraction algorithm, such as HOG~\cite{tuermer2013airborne} and SIFT~\cite{li2009robust}, to obtain feature vectors of RSI. Then, we reorganize these features and input them into an SVM classifier~\cite{bruzzone2006novel}. However, most of these algorithms have low recognition accuracy with much computational cost, which makes it difficult to meet the requirements of practical application.

\begin{figure}[tbp]
    \begin{center}
    \includegraphics[scale=1.0]{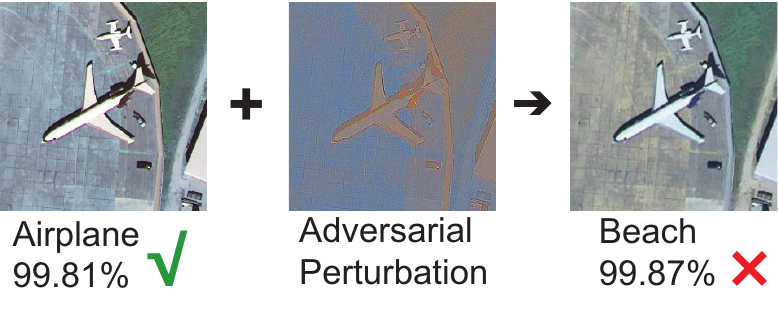}
    \end{center}
    \caption{
      On the left is the original airplane. The RSI recognition system can correctly classify it as an airplane with a confidence of 99.81\%. However, after adding an adversarial perturbation to the image, the RSI recognition system can recognize it as a beach with 99.87\% confidence. The result is wrong, and we call the modified image an adversarial example. Obviously, the adversarial example cannot affect human recognition, but it leads to system errors and has serious consequences.
    }
    \label{fig:demo1}
\end{figure}

In recent years, deep learning has achieved remarkable progress in the fields of computer vision~\cite{deng2014tutorial}. Intuitively, many excellent algorithms, especially deep convolution neural networks (CNNs), have been introduced to RSI recognition system~\cite{manno2015orientation,scott2017training,kussul2017deep}. 
Chen et al.~\cite{chen2014deep} used a single-layer AE and multi-layer SAE to learn the features of RSIs, and proposed a classification method based on extracted spatial-dominated features. This method performs better than SVM and KNN classification models. 
Chen et al.~\cite{chen2016deep} proposed a DBN-based image classification framework, which combined the spectral and spatial features of RSIs.
He used a CNN model to extract robust features from images and therefore greatly reduced the number of hyperparameters. To alleviate the over-fitting problem of CNNs, Cheng et al.~\cite{cheng2018deep} added the metric learning regular term to the CNN model by optimizing the discriminant objective function. It efficiently reduced the classification error. Zhang et al.~\cite{zhang2015scene} proposed a Gradient-Boosting Random Convolution Network which reused the weight of each CNN model and reduced parameters, making the feature extraction more efficient. Chaib et al.~\cite{chaib2017deep} used the CNN model for feature extraction and discriminant correlation analysis (DCA) for data fusion. These RSI recognition models perform much better in accuracy than traditional RSIs pattern recognition. These methods solve the problem of low accuracy and efficiency of RSI recognition, which benefits from not only the rich remote sensing data but also the increasing computing power. Importantly, these improvements are inseparable from the excellent feature extraction ability of CNNs.

However, CNNs also have many problems that are difficult to solve, especially security problems~\cite{akhtar2018threat}. Szegedy~\cite{szegedy2013intriguing} found that with the gradient algorithm, a small adversarial perturbation can be achieved on the trained model. When the input image adds this noise, the CNN model can recognize the successfully classified image as another class with high confidence. We call the modified image an adversarial example.

The problem of adversarial examples exists in various application areas of deep learning~\cite{carlini2017adversarial}. RSI recognition algorithms based on CNNs are also troubled by the problem of adversarial examples. For the classification problem of RSIs, especially those in the military field and automatic driving, it is a great security risk. The consequences can be very serious. As shown in Figure~\ref{fig:demo1}, in the military field, an attacker can generate adversarial perturbation for a target such as an airplane, and camouflage the object through a physical object. The adversarial example of the target object can easily avoid the detection of the military target detection system. This allows the originally robust target to be identified as another target with extremely high confidence. Current excellent RSI recognition systems often depend on CNNs to extract features from the target. And unlike natural images, RSIs have their unique properties, such as spectra, bands, and so on. Besides, RSIs are usually shot from a top view angle, without the foreground and back view of natural images. Therefore, it is necessary to study the characteristics of adversarial examples in RSI recognition system.

To the best of our knowledge, this paper is the first time to explore adversarial examples of RSIs. In the experiment, we trained several basic CNN models with the widest application in RSI recognition system. In these high accuracy CNNs, we use a variety of attack algorithms to generate different adversarial examples. The results of the experiment show that the problem of adversarial examples does exist in the field of RSI recognition, and high accuracy CNNs are vulnerable to adversarial examples.

Furthermore, we find that the fundamental issues on adversarial examples of RSIs: model vulnerability and attack selectivity. Model vulnerability means that different RSI recognition CNNs have different security characters. They need different costs to obtain adversarial examples. The difficulty of attack is related to the model structure and the training data. 
Attack selectivity property shows that the error classes of adversarial examples are generally concentrated. The misclassification of adversarial examples is highly similar to the correct original scene. These properties also reflect the characteristics of adversarial examples in RSI recognition.

The main contributions of this paper are:

\begin{itemize}
  \item For the first time, we analyze adversarial example problem of RSI recognition. The CNN models of RSI recognition are also vulnerable on adversarial examples. This is a new problem worth exploring.
  
  \item The vulnerability of the model to adversarial examples is also related to its model structure and the RSI dataset.
  Models with multi-scale features and datasets with rich features can effectively reduce the attack threat of adversarial examples.
  
  \item We find that adversarial example of RSI has an attack selectivity property. The result of adversarial example problem is close to the class with similar features in the CNN feature space.

\end{itemize}

The rest of the paper is organized as follows. Sections \uppercase\expandafter{\romannumeral2} introduces the research progress of adversarial examples. We explain the concept of CNN, the model structure, and the principles and algorithms of the attack algorithm in section \uppercase\expandafter{\romannumeral3}. Section \uppercase\expandafter{\romannumeral4} is the experimental setup and results analysis. In section \uppercase\expandafter{\romannumeral5}, we present conclusions and discussions.

\section{Related work}

In recent years, CNNs have been used in many tasks in the field of remote sensing, such as landmark recognition~\cite{ranjan2017hyperface}, land use classification~\cite{castelluccio2015land}, and change detection~\cite{wen2015novel}. CNNs have shown outstanding results in these areas. However, most of the research focuses on how to improve accuracy and design a good CNN structure~\cite{wu2018deepdetect,ji2018fully,hu2016edge}. There is insufficient attention to adversarial example problem of RSI recognition system. Since adversarial example was proposed by Szegedy et al.~\cite{szegedy2013intriguing}, the security of deep learning has caused widespread discussion.

Adversarial example refers to the input image formed by adding a special adversarial perturbation, which results in the model giving a false output with high confidence.
Generally, the small adversarial perturbation is not perceived by the human eye. Goodfellow~\cite{goodfellow2014explaining} pointed that adversarial examples of CNNs are due to the linear operations of the model in a high dimensional space. For feature vector operations in a high dimensional space, the small perturbations can cause great errors. Adversarial examples reduce the robustness of the model. It not only is a problem of computer vision tasks, but also exists in many fields such as natural language processing and sentiment analysis~\cite{ren2019generating}. 
We know that RSIs are high dimensional data. Intuitively, the problem of adversarial examples in RSI recognition system is unavoidable.

What's more, the attack algorithms of adversarial examples are also diverse. Goodfellow~\cite{goodfellow2014explaining} proposed the Fast Gradient Notation (FGSM) method. It first computes the gradient direction of the model loss function, and then increases the value of the loss function by adding a small adversarial perturbation in the direction. Thus, the prediction result of the model deviates from the real class of RSI. This attack algorithm is simple, but the fooling rate of the model being deceived is low. Therefore, Kurakin et al. proposed a basic iterative method (BIM)~\cite{kurakin2016adversarial}. The algorithm is an extension to the FGSM. With multiple iterations, it tries to add a noise in each step to increase the value of the loss. By each iteration, the image pixels changed are controlled within a certain neighborhood of the original picture. The BIM algorithm can effectively improve the fooling rate of attacks. Additionally, to reduce adversarial perturbation. Su et al.~\cite{carlini2017towards} proposed a single pixel attack method, which only needs to change one pixel of the original image to cause the misclassification of the classifier. The results are impressive.

Attack algorithms are also divided into multiple types. According to information acquisition by the attacker on the model, attack algorithms can be divided into a white box attack and a black box attack~\cite{akhtar2018threat}. In a white box attack, the attacker knows the type of CNN, the number of layers, and the optimization method for training. They can get the training data and even know the hyperparameters of the model. Attackers can launch more targeted attacks based on model details. Contrary to white-box attacks, attackers in black box attacks are unaware of model details. They can only use the model application background and the model's outputs to analyze the vulnerability of the model.
According to the characteristics of the attack algorithm, Biggio et al.~\cite{premlatha2018security,chakraborty2018adversarial} classify the attack scenario into three categories: evasive attack, poisoned attack and exploratory attack. The attacker in the evasive attack scenario attempts to avoid the detection of the system by constructing malicious input. This is the most common type of attack. In this scenario, the attacker cannot influence the training data of the model. The attacker in the poisoned attack scenario attempts to inject the constructed malicious data to poison the model during training process. In the exploratory attack scenario, the attacker cannot affect the training dataset or the details of the model. It acquires knowledge about learning algorithms and training data by testing the response of the model to the input data, thereby constructing effective adversarial examples.
According to whether the attack class is clear, the types of attack algorithms are also divided into targeted attacks and untargeted attacks~\cite{akhtar2018threat}. Targeted attacks attempt to make the model classify the adversarial example as a specific class, while untargeted attacks do not care about the prediction labels of adversarial examples. Untargeted attacks only require the model to misclassify the data that were originally correctly classified.

Above all, adversarial example attacks may occur during the training and testing of the model. The type of attacker is related to model details and training data. When the deep learning system exposes a lot of information to the attacker, the model can be more vulnerable.

Nowadays, RSI recognition has many CNN-based applications. They may also have adversarial example problems. These problems will have very serious consequences for the application of RSI recognition results. Moreover, RSIs have spatial and spectral properties. RSIs are different from the natural images. Therefore, it is of great significance to study adversarial example problem of RSI recognition.

\begin{figure*}[tbp]

    \begin{center}
    \includegraphics[scale=0.6]{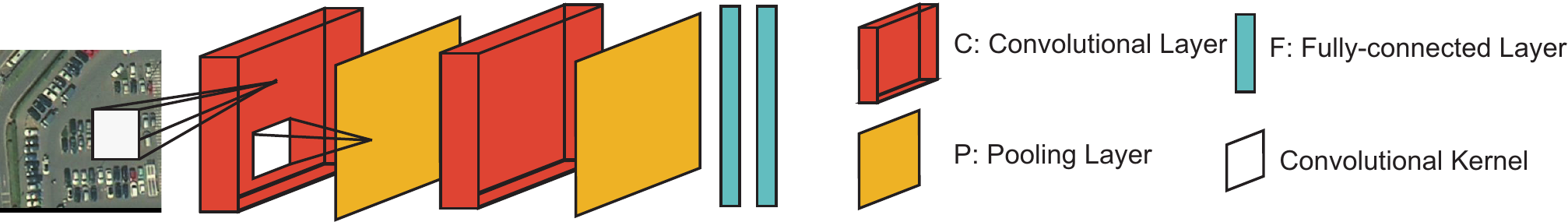}
    \end{center}
    \caption{
      The basic convolution neural network. It includes input data, alternate convolution layer and pooling layer, fully-connected layer and classifier. The input image is extracted by convolution kernel, and the final feature vector of the input image is fed to the classifier. The classifier can recognize the image.
    }
    \label{fig:cnn-base}
\end{figure*}

\begin{figure}[tbp]
    \centering
    \subfigure[Inception module]{
      \label{fig:google-m} 
      \includegraphics[scale=0.5]{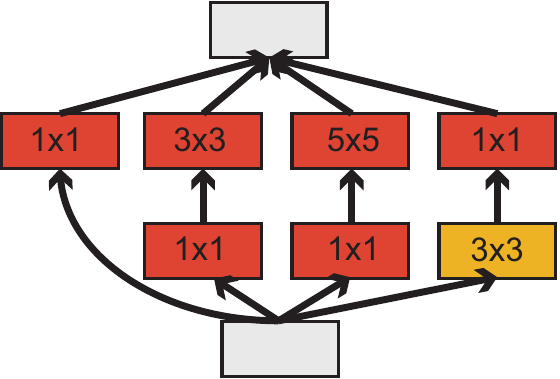}}
    \subfigure[InceptionV1]{
      \label{fig:google}
      \includegraphics[scale=0.4]{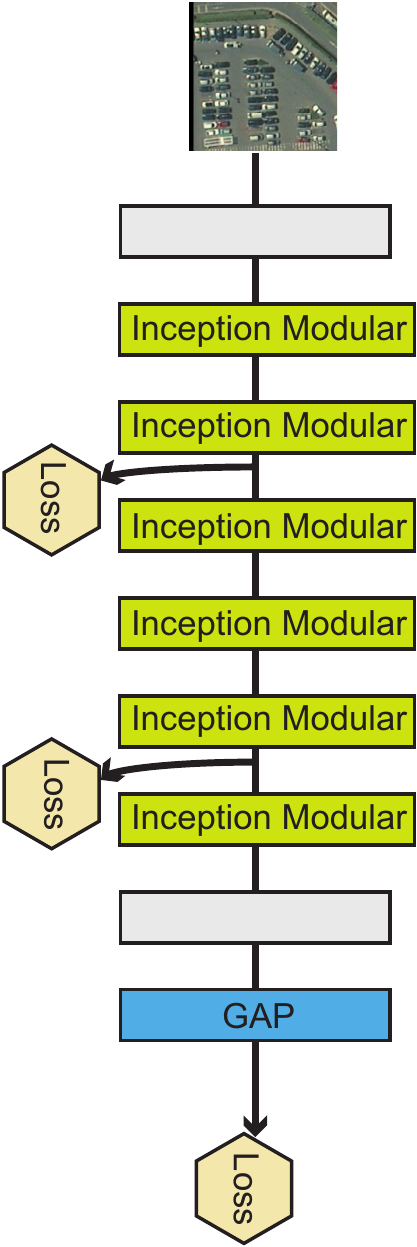}}
    \caption{
      The model structure of InceptionV1. (a) is an inception module. It is composed of several branches and the different size of convolutional kernels. The results of all branches are concatenated together in dimensions. It can extract features of different scales, which is beneficial to improve the classification accuracy. (b) is the structure of InceptionV1 model. It consists of multiple inception modules and contains three loss functions. Before the fully-connected layer, it uses the global average pooling to reduce parameters. The whole model has high accuracy, few parameters and low computation.
      }
    \label{fig:net1} 
  \end{figure}

  \begin{figure}[tbp]
    \centering
    \subfigure[Residual module]{
      \label{fig:resnet-m} 
      \includegraphics[scale=0.5]{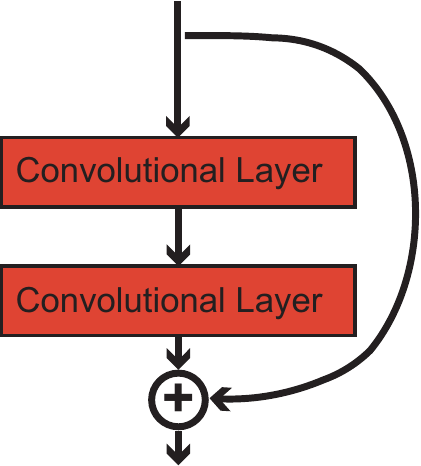}}
    \subfigure[ResNet]{
      \label{fig:resnet}
      \includegraphics[scale=0.4]{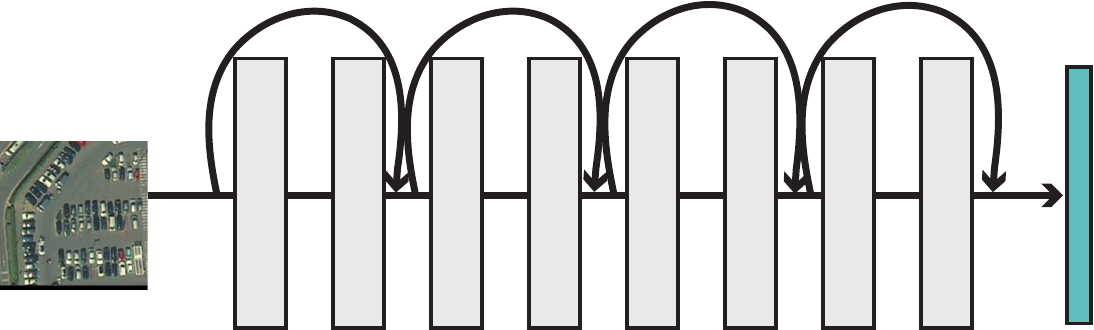}}
    \caption{   
      The model structure of ResNet. (a) is a residual module. It is implemented by a shortcut connection. The input and output of this module are added by shortcut. Without adding additional parameters, the training speed of the model can be greatly increased. (b) is the structure of ResNet model. It contains multiple residual modules. These modules make the model deeper and help solve the problems of vanishing gradient and exploding gradient.
    }
    \label{fig:net2} 
  \end{figure}

\section{Method}

This section introduces CNN briefly, and the most widely used robust CNN model structure in RSI recognition. Subsequently, we explain how to use attack algorithms to generate adversarial example of RSI.

\textbf{Definition 1:} 
We denote the input RSIs $ D = \{x^n, y^n\}$, where $x^n$ represents the input $n$ RSI, and $y^n$ represents the class of $n$ RSI. We define the RSI recognition system as $f(x)$. Generally, we can get $f(x)=y$. The system can correctly recognize the input $x$. We define $\rho$ as the adversarial perturbation. The adversarial example problem of RSIs can be represented as follows:

\begin{equation}
  \label{}
  f(x + \rho) \neq y \qquad s.t.\quad \min_{\rho} || \rho||,
\end{equation}

where $\min_{\rho} || \rho||$ denotes the regularization constraint on the adversarial perturbation. It ensures that the change of perturbation is small. After adding the adversarial perturbation, the modified RSI can cause the original correct result to be incorrect. We call this the adversarial example problem of RSI recognition.

\subsection{Convolution neural networks}
The current RSI recognition systems have better recognition capabilities, and they are mostly based on CNNs.
CNNs usually consist of convolutional layers, pooling layers, and fully-connected layers~\cite{lecun2015deep}. Each layer has a different function. There are multiple sets of alternate convolutional and pooling layers in the model, as shown in Figure~\ref{fig:cnn-base}. $C$ represents the convolutional layers, $P$ represents the pooling layer, and $F$ represents the fully-connected layer. Through these structures, RSIs are continuously reduced in dimension, and semantic features are gradually abstracted. Eventually, the model implements the classification of the RSIs. We denote $\Theta=\{ W^l,b^l\}$ to represent the parameters of the model, $W^l$ and $b^l$ to represent the weights and biases of the convolution kernel on the $l$ layer. The feedforward operation of model can be as follows,

\begin{equation}
    A^{l}(X) = f_p(f_a(W^l \ast A^{l-1}(X) + b^l)),
\end{equation}

Where $ \ast$ denotes convolutional operations. $f_a$ is an activation function, mainly including ReLu, pRelu, etc~\cite{xu2015empirical}. $f_p$ represents pooling operations, including maximum and average pooling. Pooling operations can reduce the dimension of the RSI. It cannot only reduce the computational complexity, but also improve the generalization ability of the model. $A ^{l}$ denotes the feature maps after the convolution and pooling operations.

When the feature maps are fed into the fully-connection layer, all the feature maps need to be flattened and transformed into vectors. Finally, the multi-classifier software can predict a class depending on the final outputs. When the prediction is not correct, we use cross-entropy loss function to compute the distance between the predicted results and the real labels. Then, the model can update by backpropagation algorithm~\cite{lecun2015deep}. It helps the model to reduce the loss value. The model converges gradually in the process of iteration. The optimization of the model is shown as following.

\begin{equation}
    \Theta^* = \arg \min J(\Theta; D).
\end{equation}

$J(\Theta) $ represents the empirical loss function, and $\Theta^*$ represents the weight of the model after convergence.
The trained model also performs well on the new RSI dataset and has good ability of generalization.

\subsubsection{Inception}

The outstanding model structure has made an important contribution to the success of CNN. The model structure has evolved from the AlexNet~\cite{krizhevsky2012imagenet} to several superior structures. As shown in Figure~\ref{fig:google}, InceptionV1~\cite{szegedy2015going} is the champion model of the 2014 ILSVRC classification and detection competition~\cite{ILSVRC15}, known as GoogleNet.

In the inception model structure, it introduces an inception module, as shown in Figure~\ref{fig:google-m}. Through the inception module, convolution operations are performed by multiple sets of convolution kernels of different scales, 3$\times$3 and 1$\times$1, and finally the results of different scales are stacked to form a new multi-channel data to the next inception module.

On the fully-connected layer, to reduce the parameter number, InceptionV1 also performs a new global average pooling operation (GAP). It does not need to flatten the feature maps, but instead uses a pool-like operation to turn the entire feature map into a single value. All feature maps are constructed as a feature vector to complete the remaining operations of the model. Additionally, it designs some auxiliary loss functions arranged on different layers. These losses are open during training and are closed at the time of testing. They can optimize the model. All these excellent designs not only improve the overall accuracy, but also make the InceptionV1 applied to many fields and become one of the most famous model structures. It has a very good effect on RSI recognition.

\subsubsection{ResNet}
Another model structure, ResNet~\cite{he2016deep}, won the 2015 ILSVRC competition, as shown in Figure~\ref{fig:resnet}. The model took the depth to the extreme, and for the first time the model achieved a depth of 152 layers. 
ResNet50 is the 50 layers version of ResNet.
To train deeper structures, ResNet introduced a residual module, as shown in Figure~\ref{fig:resnet-m}. It implements the delivery of information between layers through a shortcut connection. Without adding parameters, it can directly put the shallow features into the upper layer, greatly increasing the training speed of the model. ResNet has achieved a milestone on ImageNet competition~\cite{ILSVRC15}. It is the first model to surpass human recognition on ImageNet classification task. Moreover, ResNet, as the main backbone of the detection model, has achieved good results in VOC Pascal~\cite{Everingham15}, COCO~\cite{lin2014microsoft} datasets and other detection tasks~\cite{li2018detnet}. Many RSI detection applications are based on ResNet~\cite{tayara2018object}.

\begin{table*}[h]
  \caption{Details of datasets}
  \label{tab:db}
  \begin{tabular}{lllllll}
  \hline
  Datasets                               & Scene class                                                                                                                                                                                                                                                                                                                                                                                                                                                                                                                                                                                                                       & Sizes         & Total & Train & Test & Samples \\ \hline
  \multicolumn{1}{r}{UCM} & \begin{tabular}[c]{@{}l@{}} \textbf{21} (agricultural, airplane, baseballdiamond, beach,\\ buildings, chaparral, denseresidential, forest, freeway, golfcourse, harbor,\\ intersection, mediumresidential, mobilehomepark, overpass, parkinglot, river,\\ runway, sparseresidential, storagetanks and tenniscourt.)\end{tabular}                                                                                                                                                                                                                                                                                                           & 256$\times$256 & 1,680 & 1,345   & 335   & \begin{minipage}{0.1\textwidth}
      \includegraphics[scale=0.8]{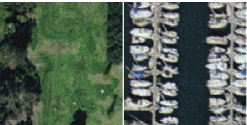}
    \end{minipage}      \\
  NWPU                          & \begin{tabular}[c]{@{}l@{}} \textbf{45} (airplane, airport, baseball diamond, basketball\\ court, beach, bridge, chaparral, church, circular farmland, cloud, commercial\\ area, dense residential, desert, forest, freeway, golf course, ground track\\ field, harbor, industrial area, intersection, island, lake, meadow, medium\\ residential, mobile home park, mountain, overpass, palace, parking lot,\\ railway, railway station, rectangular farmland, river, roundabout, runway,\\ seaice, ship, snowberg, sparse residential, stadium, storage tank, tennis\\ court, terrace, thermal power station, and wetland.)\end{tabular} & 256$\times$256 & 25,200 & 20,160   & 5,040  & \begin{minipage}{0.1\textwidth}
      \includegraphics[scale=0.8]{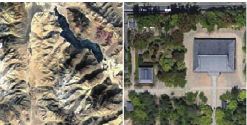}
    \end{minipage}    \\
  CLRS                                   & \begin{tabular}[c]{@{}l@{}} \textbf{20} (airport, bare-land, beach, bridge, commercial, desert, \\  farmland, forest, golf-course, highway, industrial, meadow,\\  mountain, overpass, park, parking, playground, port, railway,\\  railway-station, residential, river, runway, stadium and storage-tank.)\end{tabular}                                                                                                                                                                                                                                                                                                              & 256$\times$256 & 12,000 & 9,600   & 2,400  & \begin{minipage}{0.1\textwidth}
      \includegraphics[scale=0.8]{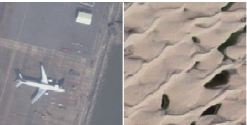}
    \end{minipage}     \\ \hline
  \end{tabular}
\end{table*}

\subsection{Attack algorithm}

These excellent models still can't escape from the problem of adversarial examples. The goal of adversarial example is mostly to make the model misclassify the minimally changed image where the original image can be classified correctly. The attack algorithms of adversarial examples are also applicable to RSIs. 

\subsubsection{Fast Gradient Sign Method}
Goodfellow~\cite{goodfellow2014explaining} developed an effective white box attack algorithm, FGSM, based on gradients. The idea is to attack the model based on the gradient direction of the model loss function. When the model converges, the weights are fixed. By computing the partial derivative of the input image through the loss function, we can obtain the optimal change direction of the input image, thereby reducing the value of loss. The FGSM algorithm does the opposite. It adds a minimal perturbation to the input image that is opposite to the optimal direction, thereby increasing the value of the loss function. It causes the model to misclassify the input image.
We denote $x$ to represents the original RSI, $x_{adv}$ to represents adversarial image of RSI by the attack algorithm. $J(\theta,x,y)$ is the model loss function, where $ \theta$ is the weights of the RSI classification model. The FGSM attack algorithm is as follows:

\begin{equation}
    X_{adv} = X + \epsilon \text{sign} (\nabla_{x} J(\theta,x,y)),
\end{equation}

$\nabla$ represents the partial derivative of the loss function, that is, the optimal direction of the input image. The sign function guarantees the same direction as the optimal change that can increases the loss value. 
Then it multiplies a small positive integer $\epsilon$, so that the model is finally misclassified. 
To constrain the adversarial perturbation of the attack algorithm, FGSM uses $L_{\infty}$ to ensure that the noise changes are small enough. The resulting $x_{adv}$ is an adversarial example of RSI $x$. 
$x_{adv}$ usually is not noticed by the human eye. It should not cause the model to change dramatically. 
However, according to the characteristics of the adversarial perturbation, the added noise increases loss value, and the model misclassifies $x_{adv}$ with high confidence. 
The FGSM attack algorithm is a change in the optimization direction, which we also call a one-step attack algorithm. The one-step attack algorithm mainly makes the loss larger to deceive the model.

\subsubsection{Basic Iterative Methods}

To increase the change of the loss, the iteratively updated multi-step attack algorithm BIM~\cite{kurakin2016adversarial} is also proposed. We know that CNN is not a linear model, and the optimal direction of change of the model will change with the input variables. Therefore, each iteration re-updates the optimization direction to get a better attack effect. BIM can be represented as follows,

\begin{equation}
    X^{i+1} = Clip\{ X^i + \epsilon \text{sign} (\nabla_{X} J(\theta,x,y)\}
\end{equation}

$i$ represents the adversarial perturbation of the $i$ iteration, and $Clip$ represents that the modified image is limited to a certain range. This attack algorithm is slower than FGSM, but it is more effective. Adversarial examples after multiple iterations are easier to deceive the model.

\subsubsection{Fooling Rate}

To evaluate the effect of attack algorithms, we introduce the definition of the fooling rate. It shows the ratio of the model's misclassified images to the attack images, which can be represented as follows

\begin{equation}
  \text{fooling rate} = \frac{\text{\# misclassification images}}{\text{\# attack images}},
\end{equation}

where $\# misclassification images$ represent the number of misclassified images after being attacked. Similarly, $\# attack images$ represent the number of images attacked. Through the fooling rate, we can evaluate the impact of the attack algorithms on model recognition under the dataset.

\section{Experiments}

In this section, we use FGSM and BIM algorithms to attack multiple converged RSI recognition models under different datasets. Then, we analyze the impact of different attack algorithms on the model.

\subsection{Experimental setup}

The attack object of adversarial example is a model with high accuracy that has been converged. In the experiment, we use three commonly used RSI datasets including: UC Merced Land Use~\cite{yang2010bag}, NWPU-RESISC45~\cite{cheng2017remote}, CLRS~\cite{lihaifeng19}. 

The UC Merced Land Use Dataset (UCM) has 21 land-use classes, each of which includes 100 256$\times$256 images. All images are extracted from the National Map Urban Area Imagery collection, including multiple towns across the United States. The resolution of the images is 1 foot.

NWPU-RESISC45 (NWPU) is a RSI scene classification benchmark data. The dataset includes 31,500 images, covering 45 classes of scenes, and each class has 700 images. The images have a resolution of 256$\times$256.

CLRS is a RSI scene classification dataset that can be used to evaluate RSI scene classification algorithms. It includes 25 classes, each of which has 600 256$\times$256 images with a total of 15,000 images.
For all datasets, we choose 80\% of them as a training set, and the remaining 20\% as a test set during the training progress.
The detail of dataset is shown in Table~\ref{tab:db}.

For deep learning models, we choose InceptionV1 and ResNet50, which are widely used in the RSI recognition system. The experimental platform is based on Ubuntu 16.04, 12$\times$3.20GHz Core i7 CPU, NVIDIA GTX 1080Ti GPU, and TensorFlow framework~\cite{tensorflow2015-whitepaper}. 
When training the model, we fine-tune the weights on three RSI training datasets based on the pre-trained InceptionV1 and Resnet50 models, and obtain 6 RSI recognition models. Their classification accuracy of the models on the training set and test set is shown in the Table~\ref{tab:nets}.

\begin{table}[tbp]
    \caption{Classification accuracy of models}
    \label{tab:nets}
    \begin{tabular}{llcc}
      \hline
      datasets & models & \multicolumn{1}{l}{train accuracy} & \multicolumn{1}{l}{test accuracy} \\ \hline
      \multirow{2}{*}{UCM} & InceptionV1 & 99.45\% & 71.73\% \\
       & ResNet50 & 100\% & 78.52\% \\ \hline
      \multirow{2}{*}{NWPU} & InceptionV1 & 99.52\% & 86.71\% \\
       & ResNet50 & 100\% & 89.62\% \\ \hline
      \multirow{2}{*}{CLRS} & InceptionV1 & 100\% & 81.17\% \\
       & ResNet50 & 100\% & 84.24\% \\ \hline
      \end{tabular}
\end{table}

The recognition results of the models sometimes are false when they are not attacked. It is meaningless to discuss adversarial example in the case of image misclassification by CNN model.
Therefore, in this experiment, only when the model classifies the original image correctly, and the classification confidence of the original image and the misclassification confidence of adversarial examples are greater than 0.7, the attack is successful.
We use the FGSM and BIM algorithms to attack this 6 RSI recognition models.

\subsection{Model vulnerability of RSIs}

Through the two attack algorithms FGSM and BIM, we find that adversarial examples are common in RSI recognition tasks. In the experiment, regardless of the training dataset and the network structures, the model vulnerability is demonstrated after the attack, that is, the image is misclassified with high confidence.

\begin{figure*}[tbp]
  \begin{center}
  \includegraphics[scale=0.7]{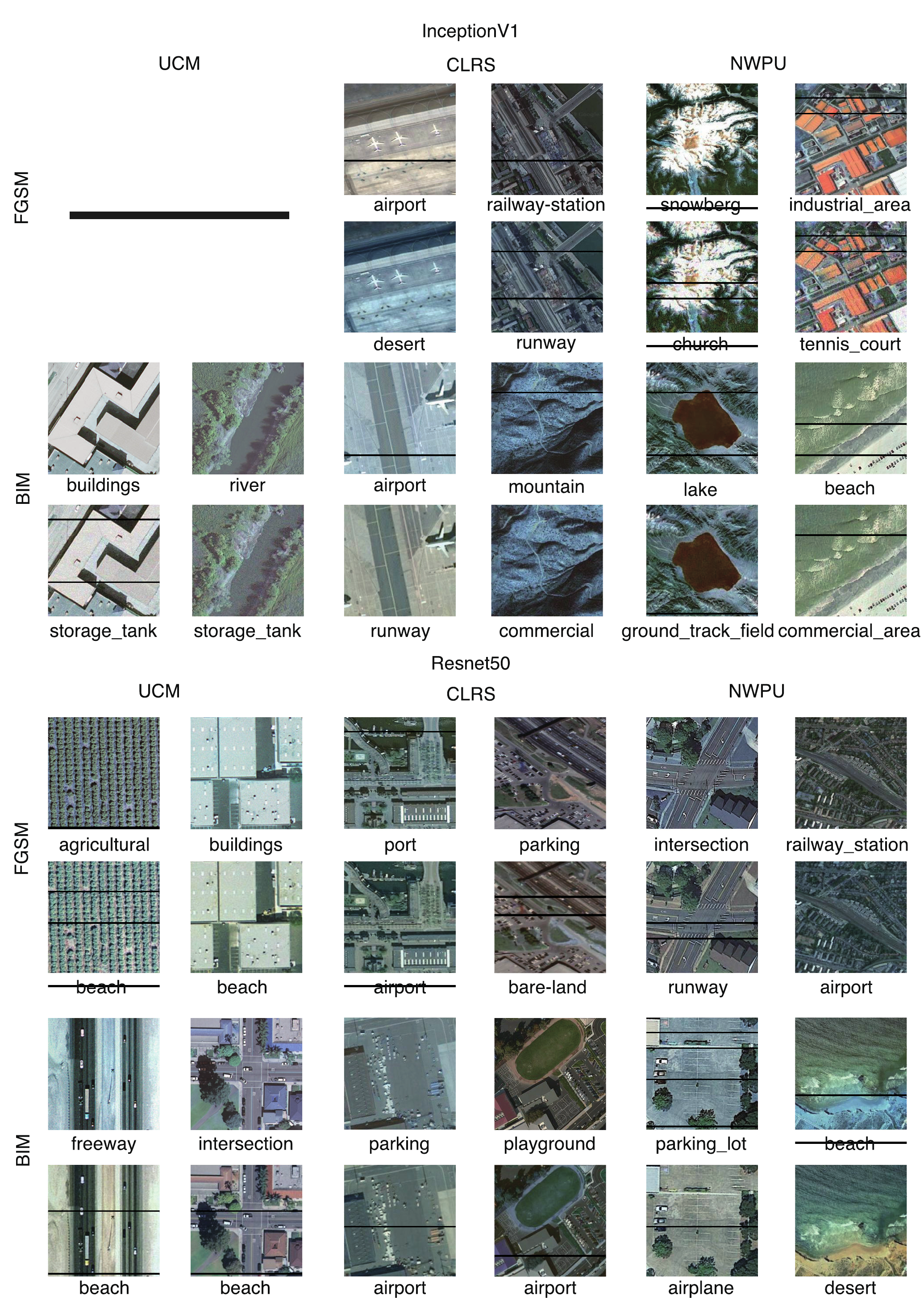}
  \end{center}
  \caption{A demonstration of adversarial examples. Under different datasets, models and attack algorithms, we show adversarial examples of attacked RSIs. Each group is represented in pairs. The upper image represents the original input image, and the lower image represents the corresponding adversarial example. We can find that the classes of the adversarial examples are misclassified.
  }
  \label{figs:ads}
\end{figure*}

The attack results are shown in Figure~\ref{figs:ads}. The original classification of RSI is correct and the confidence is more than 90\%. After the attack, we can hardly tell the difference between the adversarial example and the original input image, while RSI recognition model misclassifies adversarial examples.

Further, we find that different attack methods can bring different effects. Table~\ref{tab:fool} shows the fooling rates of 6 RSI recognition models using two attack algorithms. 

Under the two attack methods, the fooling rate of the ResNet50 model trained in the UCM dataset is higher than 80\%, and the attack fooling rate of the ResNet50 model trained by NWPU and CLRS is more than 15\%. The fooling rate of the InceptionV1 model is relatively low. The fooling rate of the BIM attack algorithm is slightly higher than that of the FGSM attack algorithm. The model vulnerability to adversarial example is multifactorial. These factors can be divided into two aspects: model vulnerability and attack characteristics.

\begin{table}[h]
    \caption{Fooling rate of attack algorithms to models}
    \label{tab:fool}
    \begin{tabular}{lllcc}
    \hline
datasets & models & \begin{tabular}[c]{@{}l@{}}attack \\ algorithms\end{tabular} & \begin{tabular}[c]{@{}l@{}}\# adversarial \\ examples\end{tabular} & \begin{tabular}[c]{@{}l@{}}fooling \\ rate\end{tabular} \\
    \hline
    \multirow{4}{*}{UCM} & \multirow{2}{*}{InceptionV1} & FGSM & 0 & 0.00\% \\
    &  & BIM & 13 & 0.77\% \\
    & \multirow{2}{*}{ResNet50} & FGSM & 1449 & 86.30\% \\
    &  & BIM & 1450 & 86.31\% \\ \hline
   \multirow{4}{*}{NWPU} & \multirow{2}{*}{InceptionV1} & FGSM & 28 & 0.37\% \\
    &  & BIM & 94 & 11.11\% \\
    & \multirow{2}{*}{ResNet50} & FGSM & 3608 & 14.31\% \\
    &  & BIM & 3608 & 14.31\% \\ \hline
   \multirow{4}{*}{CLRS} & \multirow{2}{*}{InceptionV1} & FGSM & 56 & 0.46\% \\
    &  & BIM & 130 & 1.08\% \\
    & \multirow{2}{*}{ResNet50} & FGSM & 1927 & 16.05\% \\
    &  & BIM & 1933 & 16.10\% \\ \hline
   \end{tabular}
    \end{table}

\subsubsection{Impact of the RSI model}
\begin{figure*}[tbp]
  \begin{center}
  \includegraphics[scale=1.0]{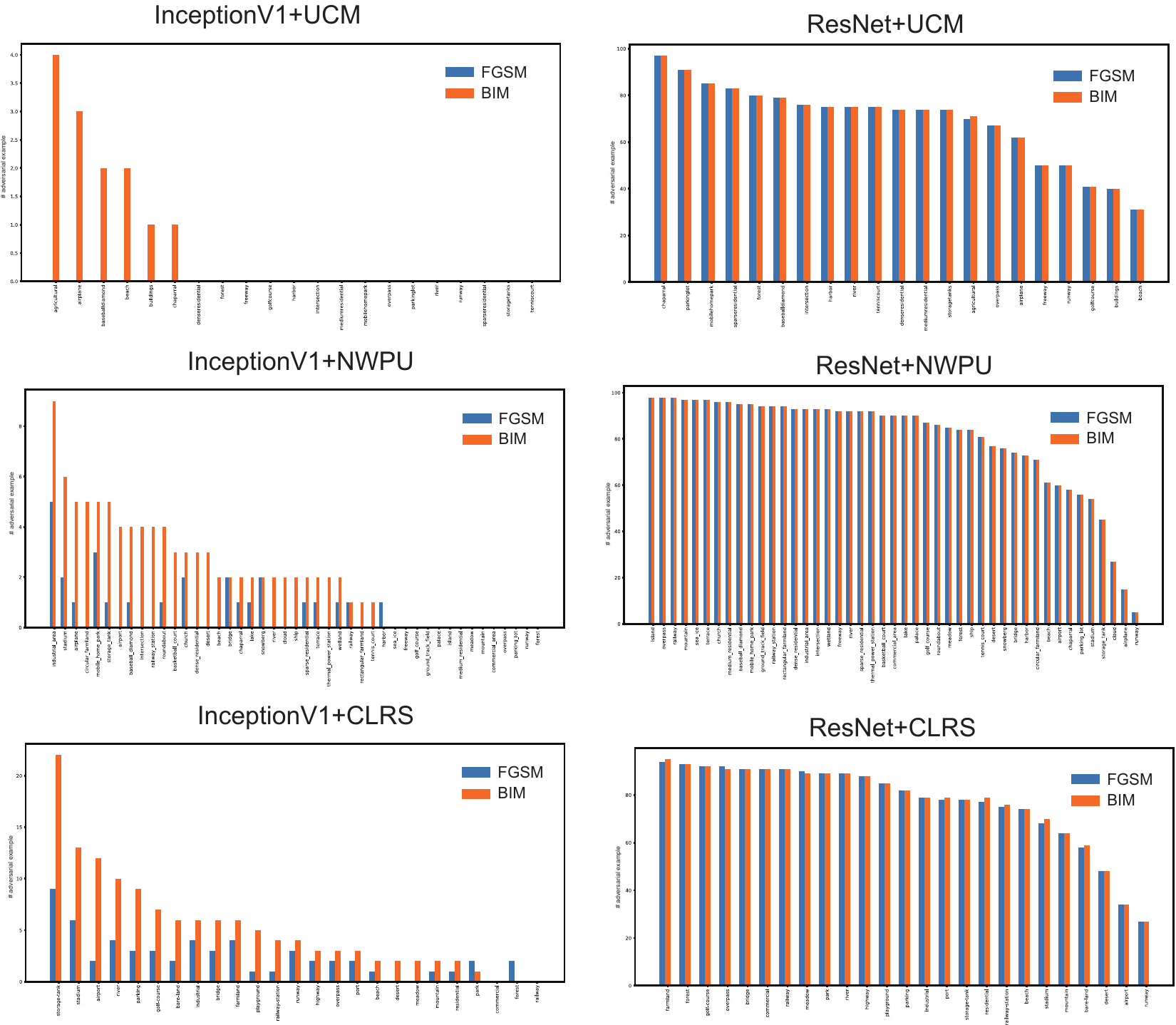}
  \end{center}
  \caption{The distribution of the model's misclassification results of adversarial examples. We find that the distribution of results is different. The results of the InceptionV1 represent a long tail distribution, while the results of ResNet50 represent a uniform distribution.
  }
  \label{fig:dis}
\end{figure*}

For model vulnerability, we believe that the RSI recognition model structure is the first factor that can affect adversarial examples. Therefore, we calculated adversarial examples obtained by all models in each class, as shown in Figure~\ref{fig:dis}.
We find that the differences in model structure have a large impact on adversarial examples. Model structure is one of the factors that influence the model vulnerability.
For the InceptionV1 and ResNet50 RSI recognition models, the attack fooling rate of the InceptionV1 model is less than 5\%, which is much lower than the attack fooling rate of the ResNet50 model.
By counting the error classes of adversarial examples after successful attack, for InceptionV1, the classes of misclassification are concentrated in few classes. The distribution of misclassified classes of adversarial examples represents a long tail distribution.
For the ResNet50 recognition model, adversarial examples are obtained in each class, and the number is relatively balanced. Only a few classes have fewer adversarial example.
We hold that the features of images in different ways of different model structures affect the model vulnerability of the RSI system.

\subsubsection{Impact of the RSI dataset}
We find that the difference of training datasets also affect the model vulnerability of RSI recognition system. From Table~\ref{tab:fool}, the fooling rates of two attack models for CNN models trained on UCM datasets are over 80\%, while the fooling rates of CNN models trained on NWPU and CLRS datasets are less than 20\%.
This is mainly determined by the number of features in the training dataset. UCM dataset has 21 classes, the total number of images is 2100. NWPU and CLRS have 45 and 25 classes, respectively, and the total numbers of images are 31.500 and 15,000, which are more substantial than UCM. When datasets have more features. These features are beneficial to defensive adversarial examples.

In high-dimensional space, small perturbations can make it difficult for features to cross the classification boundary. When training datasets involve fewer classes, their classification boundary is close to each class, and the modified images can easily lead to the features from one class to another, that is, the model is vulnerable to adversarial example. We validate this view by reducing the dimensionality of the features of the RSIs.
First, we extract the feature vectors of the last fully-connected layer of the InceptionV1 and ResNet50 models.
Then, all feature vectors are embedded into the low-dimensional space by t-SNE~\cite{maaten2008visualizing} visualization,
which is a method for mapping multi-dimensional data to two or more dimensions suitable for observation.
The close distance of data in low-dimensional space means that the similarity between data is higher.

\begin{figure*}[tbp]
    \begin{center}
    \includegraphics[scale=1.0]{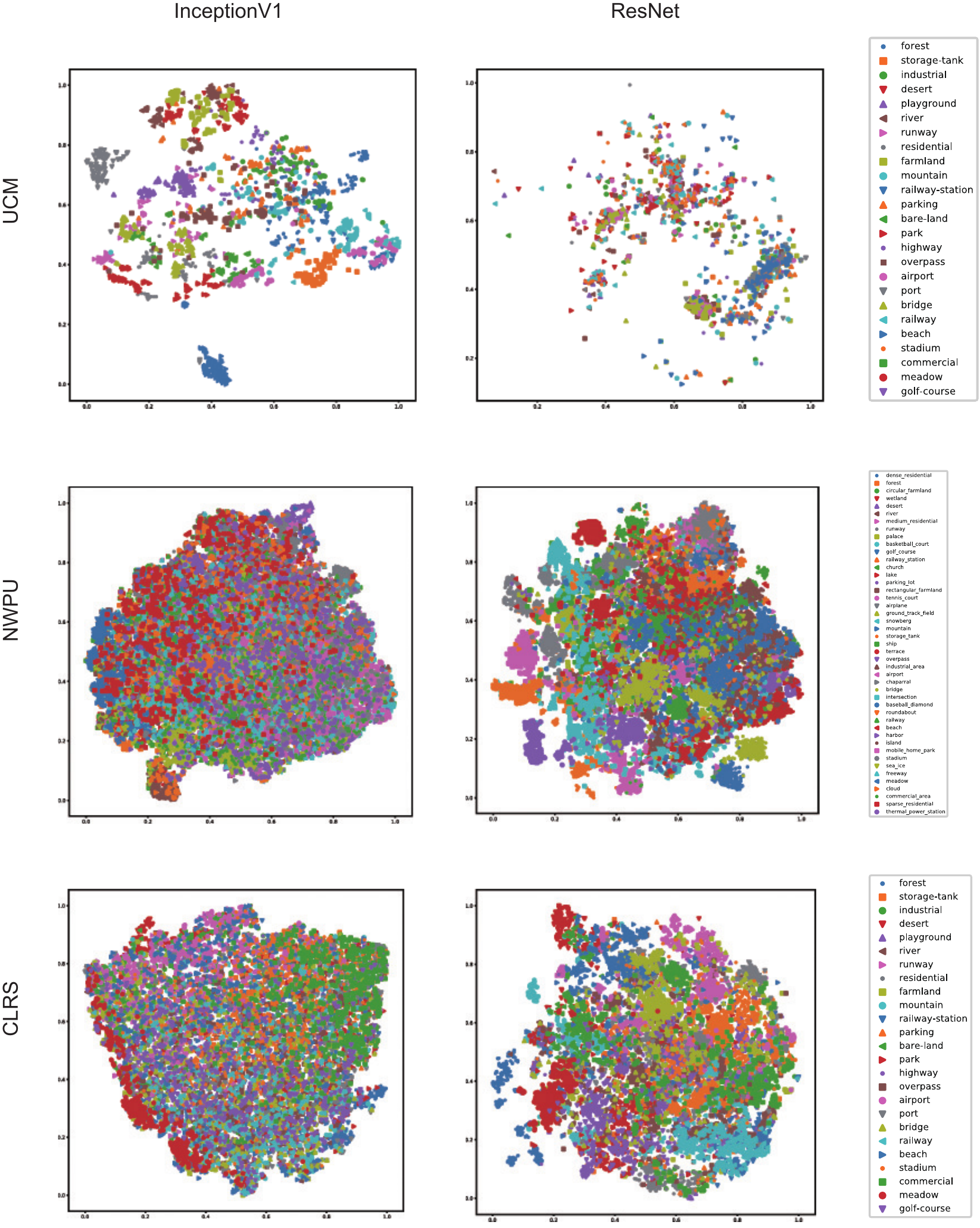}
    \end{center}
    \caption{Low-dimensional embedding of datasets under different models. }
    \label{fig:tsne}
\end{figure*}

As shown in Figure~\ref{fig:tsne}, the datasets are different in embedding space of different models. The data of UCM is sparse, and the classification distance between different features is close. On the contrary, for the NWPU and CLRS datasets, they have a mount of data, and data points mix between several classes. From Table~\ref{tab:fool}, the fooling rate on the UCM is high. This shows that the adversarial example affects the change of the data on the classification boundary. This change leads to image from one class to another. For a rich feature dataset, many features make the recognition result secure, so a small perturbation cannot affect the final recognition result. Comparing to Table~\ref{tab:fool}, we also know that the adversarial example problem on ResNet is more serious. From Figure~\ref{fig:tsne}, the classification boundaries under ResNet are distinct, but this leads to more adversarial examples. Above all, the adversarial example problem of a simple dataset is serious. The dataset is also one of the factors that affect adversarial examples.

\subsection{Attack selectivity of RSIs}
After the attack, the class of RSI object is likely to be misclassified into few specific classes by the model. We call it attack selectivity on RSI images.
Intuitively, in the high-dimensional feature space, When the FGSM and BIM attack algorithms add adversarial perturbations to RSIs, the features of adversarial example are close to the features of misclassification class.
Therefore, when the attack is successful, the classes of adversarial examples are to be in the classes of those original features in the feature space. Among all data points, points near the classification boundary are most susceptible to adversarial perturbation.

We find that RSI classes with similar features in low-dimensional space are more likely to be classes of adversarial examples when attacked.
To verify that the classes with similar features are more likely to be the classes of adversarial examples, we use the RSI features of the last fully-connected layer of the CNN model, and compute the $L_2$ distance between various RSIs and various cluster centers from the t-SNE algorithm.

\begin{figure*}[tbp]
  \centering
   \subfigure[baseball]{ 
     \includegraphics[scale=1.1]{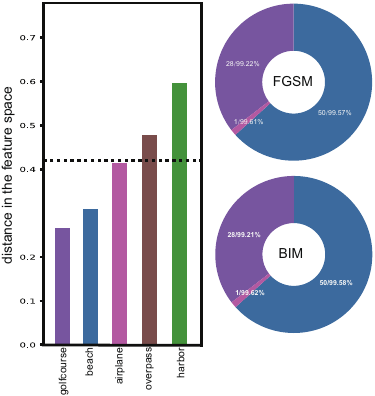}}
   \subfigure[building]{
     \includegraphics[scale=1.1]{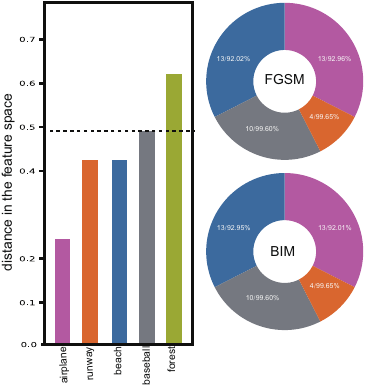}}
    \subfigure[golf course]{
      \includegraphics[scale=1.1]{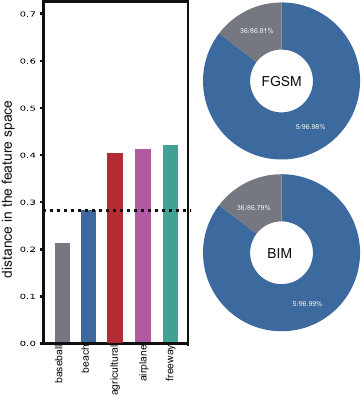}}
    \subfigure[river]{
      \includegraphics[scale=1.1]{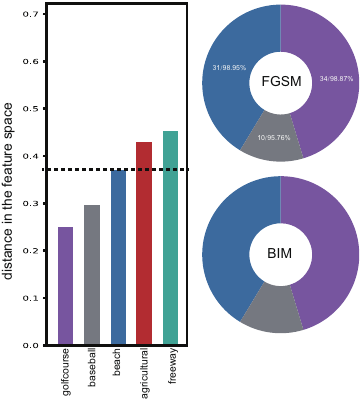}}
   \caption{A demonstration of attack selectivity, under the UCM dataset. The histogram represents the distance from the top 5 classes of other class cluster centers. The pie chart represents the distribution of adversarial examples obtained on this category. The classes of adversarial examples are below the dotted line. The adversarial classes are close to the classes in the embedding space.
   }
   \label{fig:as} 
\end{figure*}

We demonstrate several adversarial examples of ResNet50 in the UCM dataset, as shown in Figure~\ref{fig:as}. The histogram shows the top five classes in the low-dimensional embedding space. From the Figure~\ref{fig:as}, the classes of adversarial examples by the two attack algorithms are mostly concentrated in several specific classes. And these classes are among the five classes. The adversarial classes are related to the representations of the model in the feature space. The model relies on features when inferencing new RSIs. 
The adversarial perturbation changes the class of the input RSI.
On the other hand, we also constrain the changes of the adversarial perturbation so that the RSI can only move a small amount of distance. This makes the moved class close to the original class. Thus, adversarial examples of RSIs are concentrated in few classes. This characteristic is called the attack selectivity.

\subsection{Experiment discussion}

In this experiment, we verified that adversarial examples also exist in RSI recognition systems.
We find that the use of different scale training datasets and different model structures have an impact on model vulnerability.
Even the same model exhibits different model vulnerabilities for different attack algorithms.
However, when measuring the difference between the adversarial example class and the original class, we do not consider the shape of the cluster and quantitatively calculate the orientation of the RSI feature transfer, and does not discuss in detail how to defend adversarial example in RSI recognition system.

Adversarial examples of RSIs also have attack selectivity. From a geometric point of view, the RSI dataset is represented as a data point in the high-dimensional space in the CNN model, and adversarial example is deviated from the boundary of the original cluster by adding small perturbations to the data points. It causes models to misclassify them. Therefore, data points at the cluster boundary are more likely to be successfully attacked than data points near the center of the cluster.
Meanwhile, we find that most of the adversarial classes are beach. Beaches may have special characteristics.
We believe that this research can provide some ideas for the future research on the security and the defense methods on adversarial examples of the RSI recognition system.

\section{Conclusion}
In this paper, for the first time, we discuss adversarial example problem in RSI recognition. From the experiment, it shows the adversarial examples are difficult for the human eye to recognize. They allow the RSI recognition system achieve the false results. This can cause serious problems for applications based on RSI recognition systems. Through our experiment results, we propose two properties such as model vulnerability and attack selectivity to discuss adversarial example on RSI. We find that adversarial examples of RSIs are related to the similarity between classes. 
For classes with higher similarity in feature space, it is easier for them to generate corresponding adversarial examples.
This provides an important idea to further design the defense mechanism.

Adversarial examples in the field of RSI can be a problem that should be paid attention to.
In the field of RSI applications, there are still many detection and segmentation tasks, such as building detection, land cover classification, etc. The processing of these tasks is different from the image classification.
The adversarial example problem of these tasks need further study.
This is of great significance to design a secure and robust CNN-based remote sensing application system.

\bibliographystyle{IEEEtran}
\bibliography{IEEEabrv,IEEEref}

\vskip -1.5\baselineskip plus -1fil
\begin{IEEEbiography}[{\includegraphics[width=1in,height=1.25in,clip,keepaspectratio]{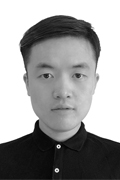}}]{Li Chen}
	received the B.Sc. and M.Sc. degree from School of Software, Central South University, china, in 2015 and 2018. He is currently pursuing the Ph.D. degree with the School of Geosciences and Info-Physics, Central South University. His research interests include geospatial trustworthy intelligence.
\end{IEEEbiography}

\vskip -1\baselineskip plus -1fil
\begin{IEEEbiography}[{\includegraphics[width=1in,height=1.25in,clip,keepaspectratio]{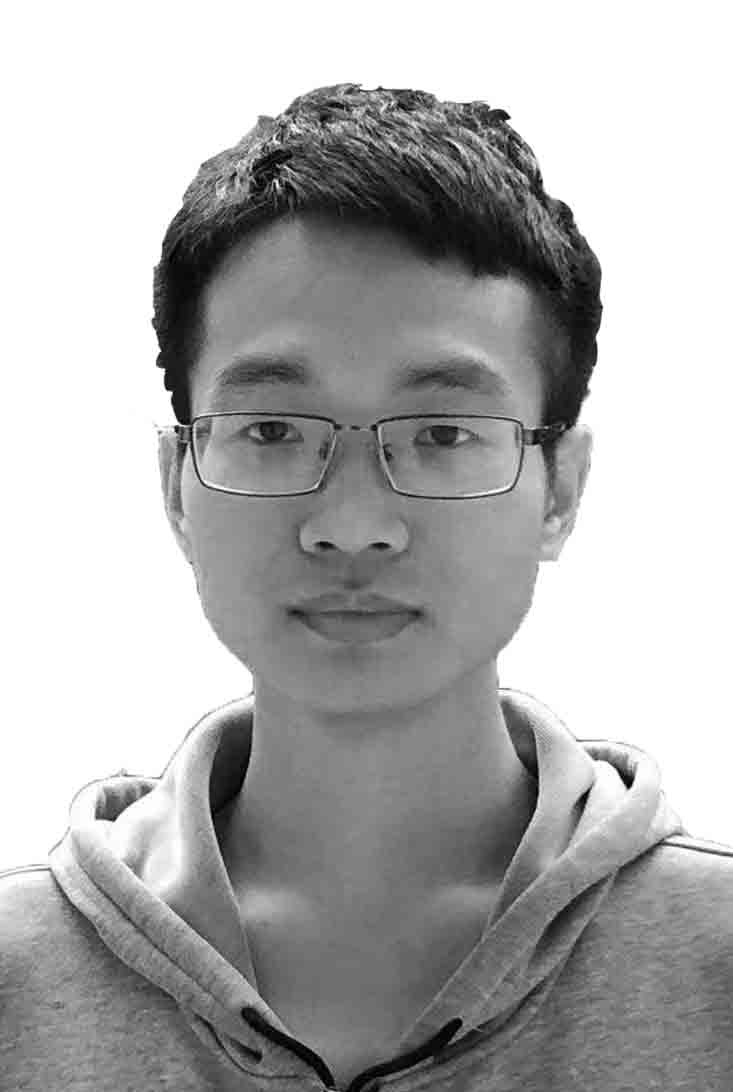}}]{Guowei Zhu}
    is currently studying at the School of Geosciences and Info-Physics, Central South University. His interests include deep learning and landmark recognition.
\end{IEEEbiography}

\vskip -1\baselineskip plus -1fil
\begin{IEEEbiography}[{\includegraphics[width=1in,height=1.25in,clip,keepaspectratio]{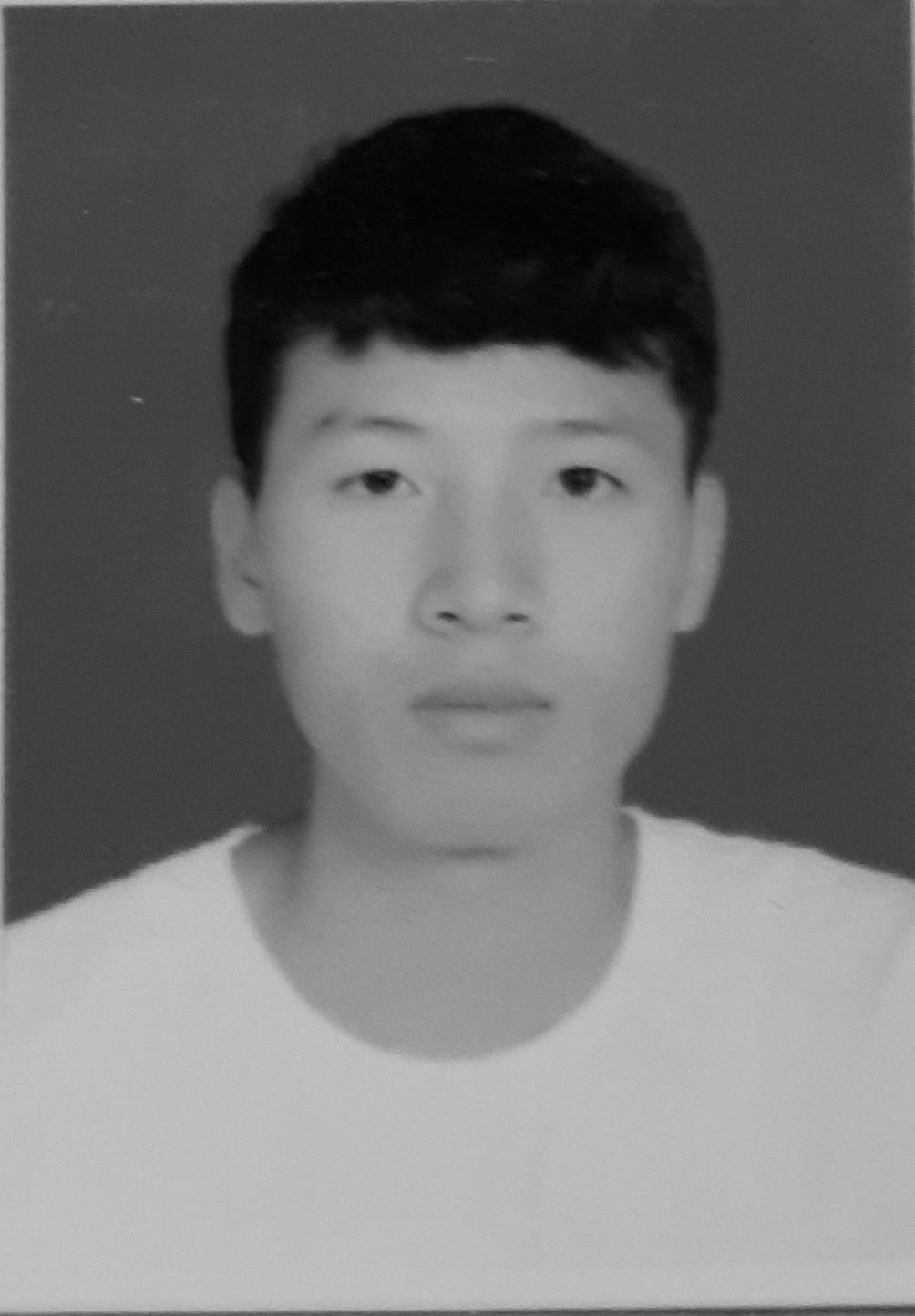}}]{Qi Li}
    is currently studying at the School of Computer Science and Engineering at Central South University. His interests include deep learning and visualization.
\end{IEEEbiography}

\vskip -1\baselineskip plus -1fil
\begin{IEEEbiography}[{\includegraphics[width=1in,height=1.25in,clip,keepaspectratio]{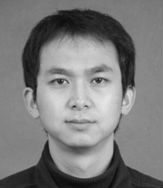}}]{Haifeng Li}
    received the master's degree in transportation engineering from the South China University of Technology, Guangzhou, China, in 2005, and the Ph.D. degree in photogrammetry and remote sensing from Wuhan University, Wuhan, China, in 2009. He is currently a Professor with the School of Geosciences and Info-Physics, Central South University, Changsha, China. He was a Research Associate with the Department of Land Surveying and Geo-Informatics, The Hong Kong Polytechnic University, Hong Kong, in 2011, and a Visiting Scholar with the University of Illinois at Urbana-Champaign, Urbana, IL, USA, from 2013 to 2014. He has authored over 30 journal papers. His current research interests include geo/remote sensing big data, machine/deep learning, and artificial/brain-inspired intelligence. He is a reviewer for many journals.
\end{IEEEbiography}

\end{document}